\documentclass{article}
\usepackage{ijcai21}

\usepackage{times}

\usepackage{soul}
\usepackage{url}
\usepackage[hidelinks]{hyperref}
\usepackage[utf8]{inputenc}
\usepackage[small]{caption}
\usepackage{graphicx}
\usepackage{amsmath}
\usepackage{booktabs}
\title{Statistical Analysis of Perspective Scores on Hate Speech Detection }

\author{
Hadi Mansourifar$^1$\and
Dana Alsagheer$^1$\and
Weidong Shi$^1$ \and
Lan Ni$^2$ \and
Yan Huang$^2$ \\
\affiliations
$^1$Computer Science Department, University of Houston\\
$^2$Valenti School of Communication, University of Houston\\

\emails
\{hmansour,dralsagh, wshi3,lni2,yhuang63\}@central.uh.edu

}

\begin{document}

\maketitle

\begin{abstract}
Hate speech detection has become a hot topic in recent years due to the exponential growth of offensive language in social media. It has proven that, state-of-the-art hate speech classifiers are efficient only when tested on the data with the same feature distribution as training data. As a consequence, model architecture plays the second role to improve the current results. In such a diverse data distribution relying on low level features is the main cause of deficiency due to natural bias in data. That's why we need to use high level features to avoid a biased judgement. In this paper, we statistically analyse the Perspective Scores and their impact on hate speech detection. We show that, different hate speech datasets are very similar when it comes to extract their Perspective Scores. Eventually, we prove that, over-sampling the Perspective Scores of a hate speech dataset can significantly improve the generalization performance when it comes to be tested on other hate speech datasets.
\end{abstract}

\section{Introduction}

Hate speech (Burnap 2014; Ross 2017) is commonly defined as any communication that disparages a person or a group on the basis of some characteristic such as race, color, ethnicity, gender, sexual orientation, nationality, religion or other characteristic (Nockleby 2000). Hate speech is a broad umbrella term for numerous kinds of insulting expressions (Schmidt 2017). One of the significant challenges in hate-speech detection on social media is to distinguish between hate speech from offensive, toxic and profane languages. That’s because there is no precise definition of hate speech in scientific community. The problem of varying definition of hate speech in different datasets makes it so hard to combine them with each other. That's why no binary hate speech model performs well on other datasets. In this paper, we address this problem using high level features extracted by Google Perspective Scores \footnote{https://www.perspectiveapi.com/}. Data preprocessing is an essential step to extract the low level features before building any model to achieve a fair and accurate result, not only applications that classify short-texts but also for clustering and anomaly detection.Text datasets contain many words and characters that do not respond well to typical methods for feature extraction which should be treated before train the data. Also, one of the most important thing to obtain fair results is that, the data we work with have no inherent biases otherwise the model will  gain  those biases and amplify them. There are multiple pre-processing approaches that can be applied to data on an automated classification task. Many researchers have made efforts to understand the leverage of different pre-processing techniques (Naseem 2020) in the automated text classification. Instead of focusing on low level features, we investigate the impact of Perspective Scores as high level features on hate speech detection. Previous work on hate speech detection investigated toxicity score of Google Perspective functions and concluded that, relying just on toxicity score would lead on a significant false positive rate. They observed that, adding curse words to normal text drastically increases the toxicity score without impacting hate speech detection probability (Fortuna 2020). However, to the best of our knowledge there is no comprehensive research in the literature to investigate the impact of high level features like severe toxicity, inflamatory, profanity, insult, obscene and spam  scores on hate speech detection. First, we assess the significance level of each mentioned scores using the statistical analysis methods like ANOVA. Afterwards, we try to investigate the significance level of possible interactions between the scores along with interpreting the coefficients and residuals. We show that, there is an interesting common pattern between different hate speech datasets in terms of Perspective Scores significance. We also discuss how the significance levels can be used as a measure to compare different datasets with each other. Finally, we train different classifiers based on Perspective Scores and test each classifier on other hate speech datasets to test the generalization capability of hate speech classifiers. Our contributions are as follows.
\begin{itemize}
    \item We analyze the significance level of Perspective Scores on hate speech detection.
    \item We prove that, over-sampling the perspective scores can improve the generalization power of of hate speech classifiers.
    \item We propose a new measure to calculate the similarity of two hate speech datasets based on significance of Perspective Scores.
\end{itemize}
\section{Related Work }
Lexicon-based approaches can gain high recall with high rates of false positives since the presence of offensive words can easily increase the toxicity and mis-classification ( Burnap and Williams 2015). Kwok and Wang discovered that,
that 86 percent of tweets categorized as hate speech just because of presence of offensive words in anti-black racism context. Given the
relatively high prevalence of offensive language and “curse words” on social media makes hate speech detection
particularly challenging (Kwok and Wang 2013). Syntactic features is the other way to handle this problem. Gitari et al. (Gitari 2015) used a rule-learning approach to extract subjective sentences. Waseem and Hovy et al. (Waseem 2016) discussed that, non-linguistic features like the gender or ethnicity can improve hate speech classification but this information is often unavailable or unreliable on social media.
Sentiment analysis is supposed to be related to hate speech detection since it is easy to assume that negative sentiment pertains to a racist messages.
That's why many researchers investigated the relatedness of hate speech and sentiment analysis using an auxiliary classifier.
some of the researchers tried to extract the high level features using text mining and web mining techniques. Djuric et  al. (Djuric 2015) developed  an  attribute-based  coding  scheme  with  eight  high-level  attributes  of  communication,  fundraising,  sharing  ideology,  propaganda  (inside), propaganda (outside), virtual community, command and control, recruitment and training. Warner et al. (Warner 2012) present a supervised  approach  that  categorizes  hate  speech  by  identifying  stereotypes  used  in  the text. Some of the identified categories include anti-Semitic, anti-Muslim and anti-African. They  create  a  language  model  based  on  the  anti-Semitic  category  and  use  correlation  to identify  the  presence  of  hate  speech  in  other  categories (Davidson 2017). Other attempts to use high level features is to use word  generalisation by extracting low-dimensional, dense vector representations via  clustering(Kumar 2020). Utilizing word embeddings is another attempt to extract high level features (Schmidt 2017). Word embedding is considered as one of the key breakthroughs of deep learning on challenging natural language processing problems. To extract the high level features in word embedding a sentence is passed to a  pretrained neural network on  a large corpus which is not  necessarily  hateful  context and the similarity to a set of selected words are calculated. Fortuna et al. (Fortuna 2020) run an empirical analysis of hate Speech datasets using the Google Perspective API and they concluded that, most of the publicly available datasets are incompatible. They showed that, even when datasets are very generic, their diverging definitions, data samples or inconsistent annotation may lead to diverging classifier performance. However, they didn't consider the Perspective Scores as high level features. Besides, empirical experiments is not enough to compare different datasets. To Address this problem, we run a comprehensive statistical analysis of Perspective Scores on hate speech detection.

\section{Statistical Analysis of Perspective Scores}
In this section, we analyze the Perspective Scores from statistical point of view. First, we extract a set of Perspective Scores as high level features for each tested dataset. Then, we analyze the significance of each score on hate speech detection task using statistical approaches like ANOVA test.
\subsection{Perspective Scores}
In order to extract high level features from hate speech instances, we use Perspective API. Perspective API was developed by Jigsaw and Google’s Counter Abuse Technology team as a part of the Conversation-AI project. The API provides several pre-trained models to compute several scores between 0 and 1 for different categories as follows (Fortuna 2020).
\begin{itemize}

\item toxicity is a “rude, disrespectful, or unreasonable
comment that is likely to make people leave a discussion.”
\item severe toxicity is a “very hateful, aggressive, disrespectful comment or otherwise very likely to make a
user leave a discussion or give up on sharing their perspective.”
\item identity attack are “negative or hateful comments targeting someone because of their identity.”
\item insult is an “insulting, inflammatory, or negative comment towards a person or a group of people.”
\item profanity are “swear words, curse words, or other obscene or profane language”
\item threat “describes an intention to inflict pain, injury, or
violence against an individual or group.”
\end{itemize}
All the trained models use Convolutional Neural Networks (CNNs), trained with GloVe word embeddings (Pennington et al., 2014) and fine-tuned during training on data from online sources such as Wikipedia and The New York Times (Fortuna 2020).
\subsection{Analysis of Variance for Regression}
Analysis of Variance (ANOVA) consists of calculations that provide information about levels of variability within a regression model and form a basis for tests of significance (Multiple Linear Regression Analysis 2018). 

% latex table generated in R 4.0.3 by xtable 1.8-4 package
% Tue May 04 23:12:10 2021
\begin{table*}[ht]
\centering
\caption{ ANOVA test results on perspective scores of BLM dataset.\\
Signif. codes:  0 ‘***’ 0.001 ‘**’ 0.01 ‘*’ 0.05 ‘.’ 0.1 ‘ ’ 1}
\begin{tabular}{lrrrrr}
  \hline
 & Df & Sum Sq & Mean Sq & F value & Pr($>$F) \\ 
  \hline
TOXICITY          & 1 & 10.34 & 10.34 & 49.20 & 8.71e-12 *** \\ 
  SEVERE\_TOXICITY   & 1 & 0.26 & 0.26 & 1.26 & 0.2629 \\ 
  IDENTITY\_ATTACK   & 1 & 1.43 & 1.43 & 6.79 & 0.0095 ** \\ 
  INSULT            & 1 & 0.02 & 0.02 & 0.10 & 0.7575 * \\ 
  PROFANITY         & 1 & 1.20 & 1.20 & 5.73 & 0.0171 \\ 
  THREAT            & 1 & 0.55 & 0.55 & 2.61 & 0.1066 \\ 
  SEXUALLY\_EXPLICIT & 1 & 0.53 & 0.53 & 2.51 & 0.1137 \\ 
  OBSCENE           & 1 & 0.78 & 0.78 & 3.72 & 0.0544 .\\ 
  SPAM              & 1 & 2.03 & 2.03 & 9.67 & 0.0020 ** \\ 
  Residuals         & 443 & 93.15 & 0.21 &  &  \\ 
   \hline
\end{tabular}

\end{table*}

\subsubsection{Test for Significance of Regression}

The test for significance of regression in the case of multiple linear regression analysis is carried out using the analysis of variance. The test is used to check if a linear statistical relationship exists between the response variable and at least one of the predictor variables. The statements for the hypotheses are: \\
$H_0: \beta_1=\beta_2=...=\beta_k=0 \\
H_1: \beta_1\neq 0 \quad \text{for at least one j} $ \\
The test for \(H_0\)  carried out using the following statistic:

%\begin{equation}
\[F_0=\frac{MS_R}{MS_E}\]
%\end{equation}

Where \(MS_R\) is the  regression mean square and \(MS_E\) is  error mean square.
If the null hypothesis \(H_0\) is true then the statistic \(F_0\) follows the \(F\) distribution with \(K\) degrees of freedom in the numerator and \(n - K+1\) degrees of freedom in the denominator. The null hypothesis \(H_0\) is rejected if the calculated statistic \(H_0\) is such that:
 
\[F_0  >  f_(a,k,n-(k+1) ) \]
where, $n$ is the total number of observations.\\

\textbf{ Calculation of the Statistic \(F_0\)}

To calculate the statistic \(F_0\) the mean squares  \(MS_R\) and \(MS_E\) must  be known. The mean squares are obtained by dividing the sum of squares by their degrees of freedom. For example, the total mean square  \(MS_T\) obtained as follows:

%\begin{equation}
%F_0= \frac {SS_T}{dof(SS_T)} 
\[MS_T=\frac{SS_T}{dof(SS_T)}  \]
%\end{equation}
where, \(SS_T\) is the total sum of squares and \(dof(SS_T)\) is the number of degrees of freedom associated with \(SS_T\). In multiple linear regression, the following equation is used to calculate \(SS_T\):

\[ SS_T = y' [I-(\frac {1}{n}J)y]\] %y
where, $n$ is the total number of observations, $y$ is the vector of observations, $I$ is the identity matrix of order $n$ and $J$ represents an  \(n \times n\) square matrix of ones. The number of degrees of freedom associated with \(SS_T\), \(dof(SS_T\)) is $n-1$. Given the  \(SS_T\) and \(dof(SS_T\)) the total mean square \(MS_T\) can be calculated.
The regression mean square \(MS_R\) is obtained by dividing the regression sum of squares \(SS_R\) by the respective degrees of freedom \(dof(SS_T\)) as follows:
\[MS_R=\frac{SS_R}{\sqrt{(SS_R)}}  \]
The regression sum of squares \(SS_R\) calculated using the following equation:
\[ SS_R =y' [H-(\frac {1}{n}J)y]\] 

where, $n$ is the total number of observations $y$ is the vector of observations $H$ is the hat matrix and $J$ represents an \(n \times n \) square matrix of ones. The number of degrees of freedom associated with \(SS_R\) and \(dof(SS_R)\), is $k$ where $k$ is the number of predictor variables in the model. Knowing the \(SS_R\) and \(dof(SS_R)\) the mean square \(MS_R\) can be calculated. The error mean square \(MS_E\) obtained by dividing the error sum of squares \(SS_E\) by the respective degrees of freedom \(SS_E\) as follows:

\[MS_E=\frac{SS_E}{dof(SS_E)}  \]
The error sum of squares, \(SS_E\), calculated using the following equation:
\[ SS_E =y' (I-H)y\]

where, $y$ is  the vector of observations, $I$ is the identity matrix of order $n$ and $H$ is the hat matrix. The number of degrees of freedom associated with \(SS_E\) and \(dof(SS_E)\) is \(n-(k+1)\) where, $n$ is the total number of observations and $k$  is the number of predictor variables in the model. Knowing \(SS_E\) and \(dof(SS_E)\) the error mean square \(MS_E\) can be calculated. The error mean square is an estimate of the variance, \( \sigma^2\)  of the random error terms, \(\epsilon_i\)
\[ \hat{\sigma}^2 =MS_E\]%need to correct 

% latex table generated in R 4.0.3 by xtable 1.8-4 package
% Tue May 04 23:26:05 2021
\begin{table*}[ht]
\caption{ANOVA test results on perspective scores of Davidson dataset.\\
Signif. codes:  0 ‘***’ 0.001 ‘**’ 0.01 ‘*’ 0.05 ‘.’ 0.1 ‘ ’ 1}
\centering
\begin{tabular}{lrrrrr}
  \hline
 & Df & Sum Sq & Mean Sq & F value & Pr($>$F) \\ 
  \hline
TOXICITY          & 1 & 10.34 & 10.34 & 49.20 & 0.00104 ** \\ 
  SEVERE\_TOXICITY   & 1 & 0.26 & 0.26 & 1.26 & 0.2629 **\\ 
  IDENTITY\_ATTACK   & 1 & 1.43 & 1.43 & 6.79 & 0.0095 ***\\ 
  INSULT            & 1 & 0.02 & 0.02 & 0.10 & 0.7575 \\ 
  PROFANITY         & 1 & 1.20 & 1.20 & 5.73 & 0.0171 **\\ 
  THREAT            & 1 & 0.55 & 0.55 & 2.61 & 0.1066 \\ 
  SEXUALLY\_EXPLICIT & 1 & 0.53 & 0.53 & 2.51 & 0.1137 \\ 
  OBSCENE           & 1 & 0.78 & 0.78 & 3.72 & 0.0544 \\ 
  SPAM              & 1 & 2.03 & 2.03 & 9.67 & 0.0020  **\\ 
  Residuals         & 443 & 93.15 & 0.21 &  &  \\ 
   \hline
\end{tabular}
\end{table*}

% latex table generated in R 4.0.3 by xtable 1.8-4 package
% Wed May 05 22:56:48 2021
\begin{table*}[ht]
\caption{ANOVA test results on interaction of BLM dataset's perspective scores. \\
Signif. codes:  0 ‘***’ 0.001 ‘**’ 0.01 ‘*’ 0.05 ‘.’ 0.1 ‘ ’ 1}
\centering
\begin{tabular}{lrrrrr}
  \hline
 & Df & Sum Sq & Mean Sq & F value & Pr($>$F) \\ 
  \hline
TOXICITY                 & 1 & 10.34 & 10.34 & 49.78 & 6.73e-12 *** \\ 
  SEVERE\_TOXICITY          & 1 & 0.26 & 0.26 & 1.27 & 0.2601 \\ 
  IDENTITY\_ATTACK          & 1 & 1.43 & 1.43 & 6.87 & 0.0091 ** \\ 
  INSULT                   & 1 & 0.02 & 0.02 & 0.10 & 0.7561 \\ 
  PROFANITY                & 1 & 1.20 & 1.20 & 5.80 & 0.0165 * \\ 
  THREAT                   & 1 & 0.55 & 0.55 & 2.65 & 0.1046 \\ 
  SEXUALLY\_EXPLICIT        & 1 & 0.53 & 0.53 & 2.54 & 0.1116 \\ 
  OBSCENE                  & 1 & 0.78 & 0.78 & 3.77 & 0.0530 .\\ 
  SPAM                     & 1 & 2.03 & 2.03 & 9.79 & 0.0019 **\\ 
  TOXICITY:IDENTITY\_ATTACK & 1 & 1.26 & 1.26 & 6.08 & 0.0140  *\\ 
  TOXICITY:SPAM            & 1 & 0.49 & 0.49 & 2.33 & 0.1273 \\ 
  TOXICITY:THREAT          & 1 & 0.03 & 0.03 & 0.13 & 0.7144 \\ 
  TOXICITY:INSULT          & 1 & 0.15 & 0.15 & 0.73 & 0.3930 \\ 
  Residuals                & 439 & 91.22 & 0.21 &  &  \\ 
   \hline
\end{tabular}
\end{table*}

\subsubsection{Test for Significance of Perspective Scores}

To evaluate the significance of Perspective Scores on hate speech detection we used ANOVA test on two datasets including BLM and Davidson datasets. First, the Perspective Score corresponding to each sentence is calculated for each dataset. Afterwards, the scores are used as predictors to fit a model given the labels of each sentence. Table 1 and Table 2 show the results which can be summarized as follows. 
\begin{itemize}
    \item In BLM dataset TOXICITY, IDENTITY\_ATTACK and SPAM are the most significant scores.
    \item In Davidson dataset TOXICITY,SEVERE\_TOXICITY, IDENTITY\_ATTACK,PROFANITY and SPAM are the most significant scores.
\end{itemize}    

\subsubsection{Test for Significance of Perspective Scores Interaction}
Also, we are interested to know if there is a significant interaction between the Perspective Scores. The interaction significance results can make it clear which one of the scores must be used together in case of filtering a dataset based on Perspective Scores. Table 3 shows the interaction significance results.
\begin{itemize}
    \item In BLM dataset there is a significant interaction between TOXICITY and IDENTITY\_ATTACK.
    \item There is no significant interaction between TOXICITY and SPAM or TOXICITY and Profanity scores.
\end{itemize} 
\subsubsection{ Quantile-Quantile plot}
The Q-Q plot, or quantile-quantile plot, is a graphical tool to help us assess if a set of data plausibly came from some theoretical distribution such as a normal or exponential (Ford 2015). A Q-Q plot is a scatter plot created by plotting two sets of quantiles against one another. If both sets of quantiles came from the same distribution, we should see the points forming a line that’s roughly straight as shown in Figure 1 which is Q-Q plot of BLM dataset. 
\begin{figure}[h]
\centering
  \includegraphics[width=60mm]{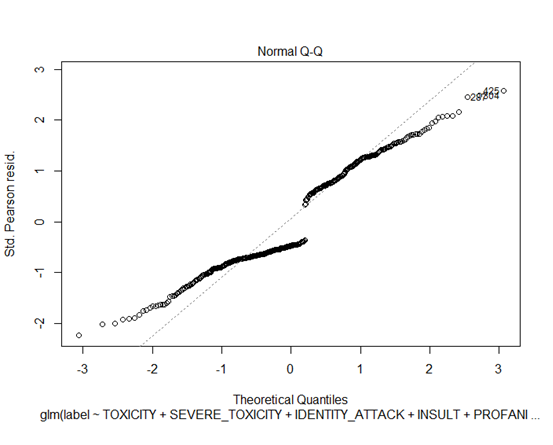}
  \caption{Q-Q plot of BLM dataset's perspective scores.}
  \label{ }
\end{figure}

\subsection{Hate Speech Classification}
The main problem in hate speech classification is that, different datasets are like isolated islands such that, a classifier trained on a dataset have poor results on another dataset. Our empirical experiments show that, imbalanced nature of hate speech datasets plays a significant role in this deficiency. To address this problem, we propose to apply over-sampling on extracted Perspective Scores as a set of high level features. First, we extract Perspective Scores of all sentences of the first dataset. Afterwards we apply over-sampling techniques like Synthetic Over-Sampling Technique (SMOTE) (Chawla 2002) and Borderline-SMOTE (Han 2002) to balance it. Finally, we train a classifier on balanced data and test it on second dataset.  Our experiments show that, over-sampling can provide a good generalization for hate speech detection task.
\subsubsection{Synthetic Over-Sampling Technique (SMOTE)}
Synthetic Minority Oversampling Technique (SMOTE) (Chawla 2002) is a method of generating new instances using existing ones from rare or minority class. SMOTE has two main steps: First, the neighborhood of each instance is defined using the k nearest neighbors of each one and Euclidean norm as the distance metric. Next, $N < k$ instances of the neighborhood are randomly chosen and used to construct new samples via interpolation (Barua 2012). Given a sample $x_{i}$ from the minority class, and N randomly chosen samples from its neighborhood $x_i^p$,  with $p = 1, . . ., N,$ a new synthetic sample $x_i^{*p}$ is obtained as follows:\\
$x_i^{*p}: = x_{i} + u (x_i^p - x_{i})$ \\
where $u$ is a randomly chosen number between 0 and 1. As a result, SMOTE works by adding any points that slightly move existing instances around its neighbors. To some extent, SMOTE is similar to random oversampling. However, it does not create the redundant instances to avoid the disadvantage of overfitting. It synthesizes a new instance by random selection and combination of existing instances.
\subsection{ Comparing Hate Speech Datasets Based on Significance of Perspective Scores}
ANOVA test on hate speech datasets is not only useful to detect the significance of each Perspective Scores but the significance results can be used to compare the distribution of different datasets. Suppose, $U$ and $V$ are two significance vector calculated by ANOVA belonging to two different hate speech datasets. The similarity of two datasets can be calculate as follows.
\[ 
Similarity (U,V)=\frac{\sum_{i=1}^{n} min(u_i,v_j)/max(u_i,v_j)}{n}
 \] 
Where, $0<Similarity (U,V) \leq 1 $ and $u_i,v_i \neq 0$
\section{Experiments}
In this section, we investigate the generalization capability of Perspective Scores. To do so, we extract the Perspective Scores of hate speech datasets. Afterwards, we train a classifier on one dataset and test that classifier on another dataset. We also empirically prove that, over-sampling can significantly improve the generalization capability of hate speech classifiers.
\subsection{Datasets}
In our experiments, we use two datasets as follows.
\begin{itemize}
    \item TweetBLM (Kumar 2020) : it includes 9165 total instances : 3084 positive and 6081 negative sentences. It has been collected by crawling Twitter data using the Tweepy which is a Python library for accessing Twitter Application Programming Interface.
    \item Davidson: it was published by (Davidson 2017). The dataset contains 24,802 tweets in English (5.77 percent labelled as hate speech, 77.43 percent as Offensive and 16.80 percent as Neither) and was published in raw text format.
\end{itemize} 

\subsection{Performance measures}
Classifier performance metrics are typically evaluated by a confusion matrix, as shown in following table. The rows are actual classes, and the columns are detected classes. TP (True Positive) is the number of correctly classified positive instances. FN (False Negative) is the number of incorrectly classified
positive instances. FP (False Positive) is the number of incorrectly classified negative instances. TN (True Negative) is the number of correctly classified negative instances. The three performance measures including precision, recall and F1 are defined by formulae (1)
through (3).
\begin{table}[h]
\begin{tabular}{|l|l|l|}
\hline
                         & \textbf{Detected Positive} & \textbf{Detected Negative} \\ \hline
\textbf{Actual Positive} & TP                         & FN                         \\ \hline
\textbf{Actual Negative} & FP                         & TN                         \\ \hline
\end{tabular}
\end{table}
\textbf{Recall} = TP/(TP+ FN), \textbf{(1)} \\
\textbf{Precision} = TP/(TP+ FP), \textbf{(2)} \\
\textbf{F1} = (2* Recall * Precision) /( Recall+ Precision) \textbf{(3)} \\

\subsection{Classification Results}
In this section, we investigate the impact of over-sampling on generalization power of hate speech classifiers. It means that, we train a classifier on dataset (A) and test it on dataset (B). To do so, first, we extract the Perspective Scores of Davidson and BLM datasets. second, we balance the davidson's Perspective Scores by SMOTE and Borderline-SMOTE. Third, we train three different base classifiers on (a) original imbalanced Davidson dataset, (b) balanced Davidson by SMOTE and (c) balanced Davidson by Borderline-SMOTE. Fourth, we evaluate trained classifiers on BLM dataset as test data. Finally, we repeat the steps one to four with different base classifiers including Decision Tree, Random Forest, Gaussian Naive Bayes, SVM, KNN and XGB. The required steps are shown in Figure 2.
\begin{figure}[h]
\centering
  \includegraphics[width=80mm]{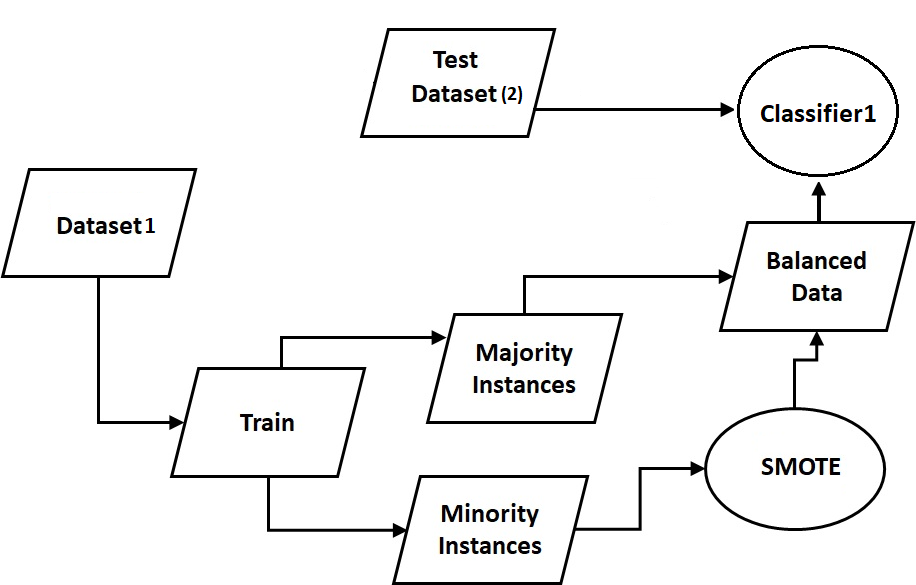}
  \caption{Over-sampling-training-test flowchart of hate speech datasets.}
  \label{ }
\end{figure}

\begin{figure*}[]
\centering
  \includegraphics[width=180mm]{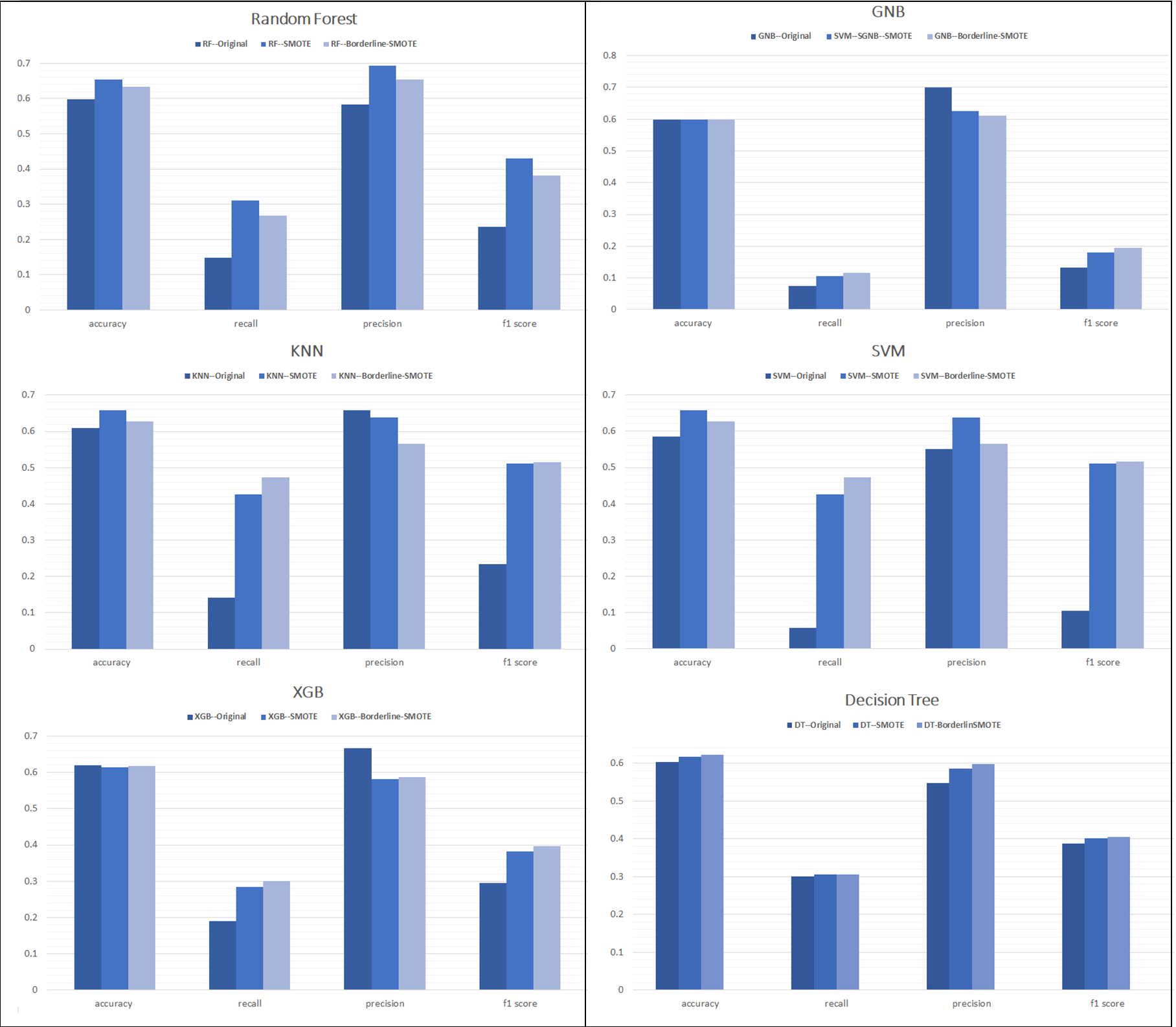}
  \caption{Classification results: The significant impact of over-sampling on generalization power of hate speech classifiers. Training data: Davidson dataset, test data: BLM dataset.}
  \label{ }
\end{figure*}

The evaluation results show that, over-sampling can significantly improve the hate speech classifiers efficiency as follows.
\begin{itemize}
\item In all base classifiers, over-sampled data can significantly improve F1-score.
\item In all base classifiers, over-sampled data can significantly improve recall score.
\item In three classifiers including DT, SVM and RF over-sampled data can improve precision.
\item In four classifiers including DT, SVM, KNN and RF over-sampled data can improve accuracy.
\end{itemize}

\section{Conclusion}
In this paper, we provided a statistical analysis on Perspective Scores and their role in hate speech detection. Besides, we proposed a novel approach to calculate the similarity of two hate speech dataset based on significance of perspective scores obtained by ANOVA test. We discussed the generalization deficiency of hate speech classifiers which means a classifier trained on a hate speech dataset can't perform well on another hate speech dataset.  To address this problem, we empirically proved that, over-sampling the perspective vectors can significantly improve the generalization power of hate speech classifiers.

\end{document}